# Atrial Fibrillation Prediction Using a Lightweight Temporal Convolutional and Selective State Space Architecture

Yongbin Lee, Ki H. Chon, *Fellow, IEEE*
*Department of Biomedical Engineering,* University of Connecticut, Storrs, United State of America
{yongbin.lee, ki.chon}@uconn.edu

*Abstract*— Atrial fibrillation (AF) is the most common arrhythmia, increasing the risk of stroke, heart failure, and other cardiovascular complications. While AF detection algorithms perform well in identifying persistent AF, early-stage progression, such as paroxysmal AF (PAF), often goes undetected due to its sudden onset and short duration. However, undetected PAF can progress into sustained AF, increasing the risk of mortality and severe complications. Early prediction of AF offers an opportunity to reduce disease progression through preventive therapies, such as catecholamine-sparing agents or beta-blockers. In this study, we propose a lightweight deep learning model using only RR Intervals (RRIs), combining a Temporal Convolutional Network (TCN) for positional encoding with Mamba, a selective state space model, to enable early prediction of AF through efficient parallel sequence modeling. In subject-wise testing results, our model achieved a sensitivity of 0.908, specificity of 0.933, F1-score of 0.930, AUROC of 0.972, and AUPRC of 0.932. Additionally, our method demonstrates high computational efficiency, with only 73.5 thousand parameters and 38.3 MFLOPs, outperforming traditional Convolutional Neural Network–Recurrent Neural Network (CNN–RNN) approaches in both accuracy and model compactness. Notably, the model can predict AF up to two hours in advance using just 30 minutes of input data, providing enough lead time for preventive interventions.

## I. Introduction

Atrial Fibrillation (AF) is the most common arrhythmia, increasing the risk of stroke, heart failure, and other cardiovascular complications [1]. Due to these risks, AF detection has been widely studied, including under challenging conditions such as the presence of premature atrial/ventricular contractions (PACs/PVCs) or motion artifacts [2], [3], [4]. Even under such conditions, recent AF detection algorithms have achieved high accuracy with lightweight models suitable for implementation on wearable devices.

While AF detection algorithms perform well in identifying persistent AF, early-stage progression, such as paroxysmal AF (PAF), often goes undetected due to its sudden onset and short duration. However, undetected PAF can progress into sustained AF, increasing the risk of mortality and severe complications. Early prediction of AF provides an opportunity to reduce disease progression through preventive therapies, such as catecholamine-sparing agents or beta-blockers [5]. These clinical concerns have led to increased research on early AF prediction to enable for timely preventive interventions.

However, AF prediction faces several key challenges: achieving high accuracy, providing sufficient lead time for preventive intervention (e.g., one to two hours in advance), and using on short input segments suitable for outpatient monitoring. Although recent studies have attempted to address these issues, they remain largely unresolved [6].

We propose a lightweight deep learning model based on a Temporal Convolutional Network (TCN) and a selective state space model known as Mamba [7], [8]. TCN utilizes causal convolutions to prevent information leakage from future time points, along with dilated convolutions and residual connections to capture long-range dependencies [7]. This architecture effectively encodes the gradual changes that occur prior to AF onset.

Mamba, a recently developed selective state space model originally introduced for natural language processing, offers advantages for time-series tasks [8]. Unlike Transformers, which use attention mechanisms but suffer from quadratic time complexity, and Recurrent Neural Networks (RNNs), which model temporal dependencies but are inherently sequential and slow to train, Mamba enables efficient parallel training with linear time complexity while maintaining strong sequence modeling capabilities.

Lastly, we used fully connected (FC) layers to predict AF. Overall, we propose a TCN–Mamba–FC architecture, in which TCN serves as a positional encoder, Mamba models temporal dynamics, and FC layers perform AF prediction.

In this study, we utilize two datasets comprising outpatient AF and NSR recordings, enabling evaluation of AF prediction performance in realistic ambulatory scenarios. Our goal is to develop a high-accuracy model capable of predicting AF two hours in advance using only 30-minute RR interval (RRI) segments, making it both practical and efficient for outpatient applications.

## II. Problem Description

Fig. 1 presents 2-hour segments of RRI signals from both AF and normal sinus rhythm (NSR) subjects. Fig. 1(a) and Fig. 1(b) show RRI recordings from two AF subjects, ranging from 2 hours before AF onset up to the onset. Fig. 1(c) displays a 2-hour RRI segment from an NSR subject. In both AF cases (Fig. 1a and 1b), ectopic beats gradually increase as the onset of AF approaches. The 30-minute window (highlighted in red) shows noticeably more frequent ectopic activity compared to earlier portions of the segment.

While Fig. 1(a) illustrates a dominant pre-AF pattern with clear ectopic irregularities, Fig. 1(b) presents a more subtle pre-AF pattern, making early prediction more difficult. The NSR example in Fig. 1(c) also contains frequent ectopic beats (highlighted in green) that resemble those seen in Fig. 1(b). This underscores the challenge of classifying between NSR and pre-AF especially more than 30 minutes prior to AF onset. To overcome this difficulty, we utilize 2-hour RRI segments from 2 hours before AF onset up to the onset itself for model training and evaluation.

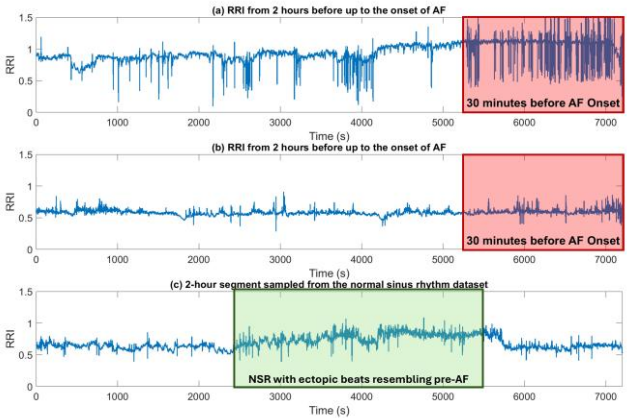

Fig. 1. Comparison of RRIs between (a), (b) pre-AF segments from the IRIDIA-AF dataset and (c) an NSR segment from the NSR RR dataset. (a) RRIs exhibit frequent ectopic beats, especially 30 minutes before AF onset. (b) Ectopic beats increase before the 30-minute mark, but pre-AF is more difficult to detect. (c) An NSR segment with frequent ectopic beats, resembling the pre-AF pattern in (b).

## III. DATASET

In this study, we used the IRIDIA-AF and NSR RR datasets, both of which contain outpatient recordings, as summarized in Table I. The NSR RR dataset comprises 24-hour RRI recordings from subjects with NSR and no documented AF episodes, although some may have a history of other cardiac conditions [9], [10]. The IRIDIA-AF dataset provides continuous ECG and RRI recordings ranging from 20 to 95 hours, with annotated AF episodes [11]. However, for the purpose of developing a lightweight prediction model, we used only the RRI data in this study.

TABLE I. THE SUMMARY OF THE DATASET

| Dataset | NSR | AF | Age | Sex (M / F) | Recording Duration |
|---|---|---|---|---|---|
| IRIDIA-AF | – | 152 | 73 ± 11 | 81 / 71 | 20 – 95 hours |
| NSR RR | 54 | – | 28 – 76 | 30 / 24 | 24 hours |

Note: NSR/AF: Number of subjects; M/F = Male/Female; Age is reported as mean ± SD or range.

For model training and evaluation, we used 2-hour pre-AF segments (2h before AF onset) from 151 AF subjects and all recordings from 54 NSR subjects. Each AF segment spans the window from 2 hours before AF onset to the onset itself. We performed a subject-wise split into training, validation, and testing sets using a 60%/20%/20% ratio, as summarized in Table II. To generate 30-minute input segments, all recordings were divided into non-overlapping 30-minute windows, resulting in a total of 604 AF segments and 2,380 NSR segments.

To ensure the robustness of our results, we conducted five

TABLE II. NUMBER OF SUBJECTS (SEGMENTS) USED FOR TRAINING, VALIDATION, AND TESTING.

| Dataset | Training | Validation | Testing |
|---|---|---|---|
| IRIDIA-AF | 90 (352 – 354) | 30 (120 – 121) | 31 (121 – 122) |
| NSR RR | 32 (353) | 11 (121) | 11 (121) |

Note: Dataset split by subjects (number of segments).

separate train/validation/test splits using different random seeds, where the AF and NSR datasets were independently and randomly partitioned into training (60%), validation (20%), and testing (20%) sets. This resulted in slightly different segment distributions across the splits, particularly for AF subjects who had multiple AF episodes. We trained and evaluated the model independently on each of the five splits and reported the mean of the evaluation metrics to account for variability due to subject-wise sampling.

## IV. METHOD

### A. AF prediction model Architecture

Fig. 2 summarizes our proposed AF prediction model architecture. The model begins with a TCN block, as illustrated in Fig. 2(b). The input RRI signal, with a dimension of (1, 1800), is first passed through a causal dilated 1D convolutional layer (Conv1D) with a kernel size of 3, expanding the channel dimension to 32. This is followed by Batch Normalization (BN), ReLU activation, and a Dropout layer (rate = 0.2) to stabilize training and reduce overfitting. Next, a second causal dilated Conv1D layer and another BN layer are applied. A skip connection adds the TCN block input, which is passed through a 1×1 Conv1D and BN layer, to the main path. This merged output is then followed by a final ReLU activation. The second and third TCN layers follow the same structure, but with fixed input/output channels of 32 and increasing dilation rates of 2 and 4, respectively. After the three residual TCN blocks, a MaxPooling1D layer reduces the temporal resolution, compressing the feature map from a dimension of (32, 1800) to (32, 900).

Next, the feature map is passed to the Mamba layer, as illustrated in Fig. 2(c). The input feature map is processed by the Mamba Selective State Space Model, which captures sequential dependencies within the encoded features. A residual connection is then added to the output, followed by Layer Normalization. After this, a feed-forward network (FFN) is applied. This FFN, shown in Fig. 2(d), is convolution-based and is particularly effective for capturing temporal dynamics, as neighboring hidden states tend to be more closely related [12]. Another residual connection is merged with the FFN output, followed by Layer Normalization and Dropout (rate = 0.2).

Finally, we apply Global Average Pooling (GAP) and Global Max Pooling (GMP) and concatenate their outputs to form a feature vector of dimension 64. This vector is then passed through two Fully Connected (FC) blocks, as shown in Fig. 2(e). Each FC block consists of an FC layer, BN, and ReLU activation. The first FC block reduces the dimensionality from 64 to 32, and the second reduces it to 2. The final output is obtained using a SoftMax activation function to predict AF probability.

### B. Performance Evaluaation

For performance evaluation, we computed five key metrics: sensitivity (Sens.), specificity (Spec.), weighted F1-score (F1), area under the receiver operating characteristic curve (AUROC), and area under the precision-recall curve (AUPRC). Accuracy was excluded from the evaluation due to class imbalance in the datasets.

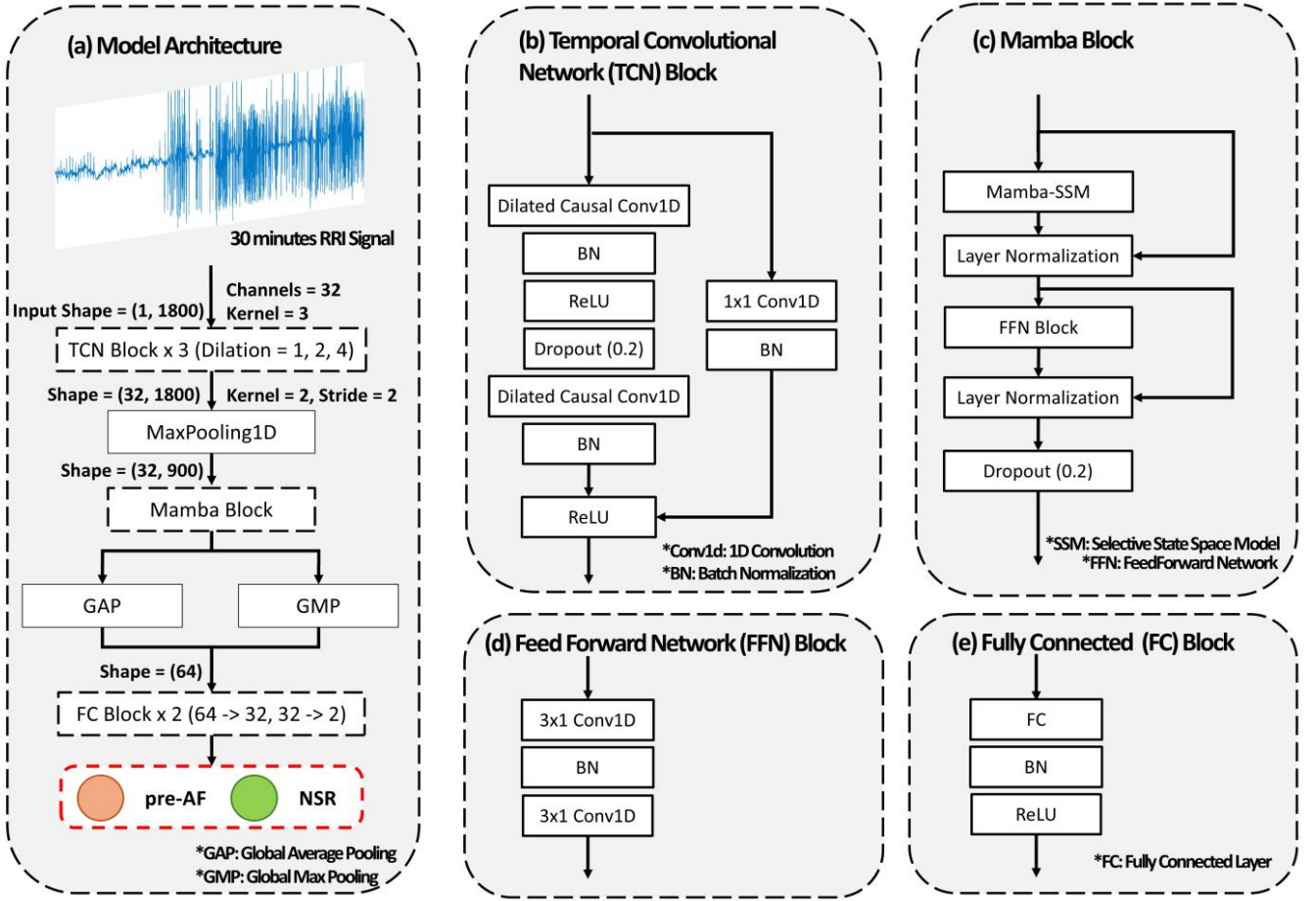

Fig. 2. Our proposed AF model architecture

*C. Optimization*

All models were trained using a batch size of 16, the cross-entropy loss function, and the AdamW optimizer with a learning rate of 1e−4 and weight decay of 1e−4. Training was conducted for up to 1000 epochs, with early stopping applied based on validation loss, using a patience of 10 epochs to select the best-performing model.

Experiments were conducted on Google Colab Pro using an NVIDIA A100 GPU (40 GB VRAM). The software environment included Python 3.11 and PyTorch 2.4.0 with CUDA 12.1. Additional libraries used included causal-conv1d (v1.4.0) and mamba-ssm (v2.2.2), installed via pip.

## V. RESULTS

*A. Performance Comparison with Prior Models*

To compare the performance of our model with existing deep learning approaches, we re-implemented the reference models based on the architecture described in the original publications and trained them on our dataset using the same training, validation, and testing strategy, along with the same optimization settings described in Section IV-C.

Table III presents a comparative summary of the results. Grégoire et al. [13] proposed an AF prediction model using RRIs, comprising a convolutional neural network (CNN) followed by a bidirectional gated recurrent unit (Bi-GRU).

TABLE III. PERFORMANCE METRICS OF OUR MODEL COMPARED TO PREVIOUS DEEP LEARNING APPROACHES

| Model | Sens. | Spec. | F1 | AUROC | AUPRC |
|---|---|---|---|---|---|
| Grégoire et al. [13] | 0.839 | 0.867 | 0.869 | 0.909 | 0.788 |
| Lin et al. [9] | **0.930** | 0.892 | 0.905 | 0.965 | 0.907 |
| Hannun et al. [14] | 0.813 | 0.884 | 0.876 | 0.900 | 0.755 |
| **Our model** | 0.908 | **0.933** | **0.930** | **0.972** | **0.932** |

Lin et al. [9] developed a model using RRIs with a CNN and bidirectional long short-term memory (Bi-LSTM). Although their original implementation included demographic features such as age and sex, we excluded these variables in our replication to focus on physiological signals. Hannun et al. [14] introduced a deep residual CNN architecture composed of 15 residual blocks for arrhythmia classification, including AF, using ambulatory ECG signals.

As shown in Table III, our model achieved superior performance in specificity (0.933), F1-score (0.930), AUROC (0.972), and AUPRC (0.932). While the model by Lin et al. [9] achieved the highest sensitivity, our model outperformed all others across the remaining metrics, indicating a more balanced and robust overall performance.

*B. Performance Comparison in Model Complexity*

Table IV presents a comparison of model complexity across four deep learning models. The metrics used are the number of total trainable parameters and the number of floating-point operations (FLOPs), which reflect both memory and computational costs, respectively. Among the models, the Hannun et al. [14] model is the most computationally expensive, with 10.7 million parameters and 517 million FLOPs, as it employs a deep residual network with 15 CNN residual blocks. The Lin et al. [9] and Grégoire et al. [13] models show moderate complexity, with 300K and 200K parameters, and FLOPs of 55.4M and 54.4M, respectively. In contrast, our proposed TCN–Mamba model achieves significantly lower complexity, with only 73.5K parameters and 38.3M FLOPs. This highlights the efficiency of our architecture, making it suitable for deployment in resource-constrained settings such as wearable or mobile devices.

TABLE IV. MODEL COMPLEXITY COMPARISON IN TERMS OF PARAMETERS AND COMPUTATIONAL COST

| Model | Total Parameters | FLOPs |
|---|---|---|
| Gregoire [1] | 200K | 54.4M |
| Lin [2] | 300K | 55.4M |
| Hannun [3] | 10.7M | 517M |
| **Our model** | **73.5K** | **38.3M** |

## VI. DISCUSSION

This study presents a lightweight and computationally efficient model for AF prediction 2 hour in advance using only RRIs, integrating a Temporal Convolutional Network (TCN) for positional encoding and the Mamba Selective State Space Model for sequence modeling. Our model demonstrates high predictive performance, achieving a sensitivity of 0.908, specificity of 0.933, F1-score of 0.930, AUROC of 0.972, and AUPRC of 0.932 in subject-wise testing.

Compared to previously proposed CNN–RNN architectures, such as those by Grégoire et al. [13], Lin et al. [9], and Hannun et al. [14], our approach outperformed most evaluation metrics while using significantly fewer parameters (73.5K) and lower computational cost (38.3 MFLOPs). These results suggest that TCN and Mamba are a promising alternative to more resource-intensive RNN-based models.

A key advantage of our approach is its ability to predict AF up to two hours before onset using only 30-minute RRI segments. This is especially critical for identifying paroxysmal AF, which is challenging to detect due to its sudden and short-term nature. It enables timely preventive interventions (e.g., beta-blockers or clinical alerts), which may help prevent the progress of AF.

While our model performs well in metrics, further validation across larger and more diverse outpatient datasets is necessary to assess its generalizability. Additionally, the reliance on RRI alone, while advantageous for wearable devices, may limit interpretability compared to multi-modal approaches.

## VII. CONCLUSION

We proposed a novel deep learning model for early AF prediction that integrates a Temporal Convolutional Network (TCN) and the Mamba Selective State Space Model to extract and model temporal features from RRIs. The model demonstrates high predictive accuracy while maintaining a small parameter size and low computational overhead, making it suitable for real-time applications. Comparing to traditional CNN–RNN models, our TCN–Mamba architecture provides both computational efficiency and superior performance, particularly in F1-score, AUROC, and AUPRC. The ability to predict AF two hours in advance using only 30-minute input windows offers a promising step toward proactive AF monitoring and prevention. In future work, we aim to incorporate additional biosignals such as ECG, PPG, EDA, and evaluate performance across multiple cohorts to enhance model robustness and clinical utility.